\documentclass{article}

\usepackage[]{iclr2024_conference}

\usepackage[utf8]{inputenc} %
\usepackage[T1]{fontenc}    %
\usepackage{hyperref}       %
\usepackage{url}            %
\usepackage{booktabs}       %
\usepackage{amsfonts}       %
\usepackage{nicefrac}       %
\usepackage{microtype}      %
\usepackage{xcolor}         %
\usepackage{graphicx}       %
\usepackage{caption}
\usepackage{subcaption}
\usepackage{amsmath}
\usepackage{adjustbox}
\usepackage{hyphenat}
\usepackage{tikz}
\usepackage{wrapfig}

\usetikzlibrary{svg.path}

\usepackage{enumitem}

\usepackage[colorinlistoftodos,prependcaption,textsize=tiny]{todonotes}
\newcommand{\orcid}[1]{\href{https://orcid.org/#1}{\textcolor[HTML]{A6CE39}{XXX}}}

\definecolor{orcidlogocol}{HTML}{A6CE39}
\tikzset{
  orcidlogo/.pic={
    \fill[orcidlogocol] svg{M256,128c0,70.7-57.3,128-128,128C57.3,256,0,198.7,0,128C0,57.3,57.3,0,128,0C198.7,0,256,57.3,256,128z};
    \fill[white] svg{M86.3,186.2H70.9V79.1h15.4v48.4V186.2z}
                 svg{M108.9,79.1h41.6c39.6,0,57,28.3,57,53.6c0,27.5-21.5,53.6-56.8,53.6h-41.8V79.1z M124.3,172.4h24.5c34.9,0,42.9-26.5,42.9-39.7c0-21.5-13.7-39.7-43.7-39.7h-23.7V172.4z}
                 svg{M88.7,56.8c0,5.5-4.5,10.1-10.1,10.1c-5.6,0-10.1-4.6-10.1-10.1c0-5.6,4.5-10.1,10.1-10.1C84.2,46.7,88.7,51.3,88.7,56.8z};
  }
}

\newcommand\orcidicon[1]{\href{https://orcid.org/#1}{\mbox{\scalerel*{
\begin{tikzpicture}[xscale=0.1, yscale=-0.1,transform shape]
\pic{orcidlogo};
\end{tikzpicture}
}{|}}}}

\newcommand{\comment}[1]{}  %

\iclrfinaltrue

\title{\textbf{\nohyphens{Generalized Policy Learning for Smart Grids: FL TRPO Approach}}}

\author{Yunxiang Li \\
MBZUAI, UAE \\
\texttt{yunxiang.li@mbzuai.ac.ae} \\ 
\And
Nicolas M. Cuadrado \\
MBZUAI, UAE \\
\texttt{nicolas.avila@mbzuai.ac.ae} \\ 
\And
Samuel Horv\'ath \\
MBZUAI, UAE\\
\texttt{samuel.horvath@mbzuai.ac.ae}$\qquad\qquad\quad\qquad\qquad $
\And
Martin Taká\v{c} \\
MBZUAI, UAE\\
\texttt{martin.takac@mbzuai.ac.ae}
}

\usepackage{array}
\usepackage{arydshln}
\begin{document}

\maketitle

\begin{abstract}

The smart grid domain requires bolstering the capabilities of existing energy management systems; Federated Learning (FL) aligns with this goal as it demonstrates a remarkable ability to train models on heterogeneous datasets while maintaining data privacy, making it suitable for smart grid applications, which often involve disparate data distributions and interdependencies among features that hinder the suitability of linear models. This paper introduces a framework that combines FL with a Trust Region Policy Optimization (FL TRPO) aiming to reduce energy-associated emissions and costs. Our approach reveals latent interconnections and employs personalized encoding methods to capture unique insights, understanding the relationships between features and optimal strategies, allowing our model to generalize to previously unseen data. Experimental results validate the robustness of our approach, affirming its proficiency in effectively learning policy models for smart grid challenges.

\end{abstract}

\section{Introduction}

The world faces the pressing challenge of climate change mainly due to energy use, leading to an increased focus on renewable energy sources and energy-efficient systems as a mitigation technique. The smart grid domain considers energy storage systems, cutting-edge technologies, distributed energy resources management, and automated demand response as promising solutions that can help optimize energy usage and reduce carbon emissions. However, the energy grid is a complex and interconnected system that suffers from the constantly changing scenarios brought by the random nature of renewable.

Machine Learning seems to be a pivotal tool to address the complexities of the scenario mentioned above \cite{DBLP:journals/corr/abs-1906-05433}. Still, real-life applications must address the generalization of models and preservation of sensitive energy usage data. We propose an FL TRPO model. This framework harnesses inherent generalization capability and privacy guarantees from FL to discern latent relationships within the feature space. In conjunction with TRPO, strategically incorporating a purposefully designed model to leverage prior knowledge regarding the features, the FL TRPO model exhibits the robustness necessary to map from feature space to optimal policy outcomes effectively.

\begin{itemize}[left=0pt,noitemsep,nolistsep]
    \item This research evaluates the performance of FL TRPO in the context of a smart grid problem, leveraging the faster convergence of TRPO with the fewer communication rounds time of FL.
    \item FL generalize effectively to previously unseen datasets \cite{DBLP:journals/corr/abs-2212-08343}. We constructed training and testing datasets with different distributions to showcase this ability in the smart grid scenario. We incorporated intricate environmental relationships where the reward is a non-linear function of the features.
    \item By exposing agents to varying training and testing data distributions, we conducted a comparative analysis of performance with and without FL.
    \item Harnessing our prior understanding of the features, we engineered a model capable of computing optimal policies, thereby elevating the overall model performance.
\end{itemize}

\section{Related Work}
{\bf Federated Learning.}
FL provides methods to train models across decentralized agents while keeping privacy. Federated Averaging (FedAvg) \cite{DBLP:conf/aistats/McMahanMRHA17} stands as the pioneering framework in FL, where agents exchange weights after multiple updates on their respective local data. Researchers have enhanced FL by building upon the notion of training decentralized agents. Gossip algorithms ensure efficient communication \cite{DBLP:conf/icml/KoloskovaSJ19}, split learning offers a method to achieve personalization \cite{DBLP:journals/corr/abs-1812-00564}, differential privacy ensures to safeguard privacy \cite{DBLP:journals/corr/abs-1712-07557}, novel computation and communication techniques have been developed to handle heterogeneous data \cite{DBLP:conf/iclr/Diao0T21}, and analysis generalization especially in FL \cite{DBLP:conf/iclr/0002MNS22}. Benefiting from its ability to handle heterogeneous data training, preserving privacy, and generalization on unseen data, FL demonstrates promising performance in addressing challenges related to smart grid scenarios.

{\bf Smart grids problem.}
Recently, the challenges within smart grid systems have garnered attention across academic and industrial spheres. \cite{DBLP:journals/corr/abs-2104-09785} devised a model-based reinforcement learning (RL) approach tailored to multi-energy systems, enhancing it with constrained optimization algorithms to derive optimal solutions. Meanwhile, \cite{DBLP:journals/tii/SuWLZLCC22} introduced a Federated RL (FRL) methodology, considering the dynamics between Energy Data Owners (EDOs) and Energy Service Providers (ESPs), addressing free-rider phenomena and non-independent and identically distributed (IID) data distribution nuances. In \cite{DBLP:conf/camad/RezazadehB22}, this work was further enriched by harnessing a discretized Soft Actor-Critic (SAC) framework within the FRL paradigm, leading to faster convergence and mitigating privacy breaches. Similarly, \cite{DBLP:journals/tii/LeeC22} leveraged FRL and demonstrated faster convergence and better performance than agents without FRL. Furthermore, \cite{DBLP:conf/iclr/CuadradoGT23} introduced a hierarchical FRL architecture, yielding performance enhancements. \cite{DBLP:journals/corr/abs-2303-08447} presented a holistic framework for smart grid problems with three layers of different objectives. Despite these pivotal advances highlighting FRL's potency in energy management difficulties, the focus has primarily centered on its operational efficacy. 
\section{Problem and Environment}

We based our environment setting on CityLearn \cite{DBLP:conf/sensys/Vazquez-Canteli19}, which provides an OpenAI Gym environment for building energy coordination and demand response based on a real dataset \cite{oedi_4520}. We generated two environments (training and evaluation) containing five buildings, each equipped with a non-shiftable load, solar panels, and a controllable battery managed by the agent. The grid price pattern mirrors demand fluctuations and is charged as a Time-of-Use (ToU) electricity rate, peaking during high-demand hours, while the solar panels contribute energy throughout the daylight hours. Allowing the agent to stockpile energy during off-peak times and, when solar generation exceeds immediate needs, later release stored energy during high-price periods. Emissions follow the actual emission rate from the grid mix (kgCO2e/kWh) for the dataset used in the environment.

Each episode spans twenty-four hours, and $t$ is the index for the time step. For a building $i$, an agent considers the solar panels' energy and the grid energy price to decide the charge or discharge of the battery in a time step ($E^{\text{batt}}_{t,i}$) aiming to meet the building's load requirements. Its primary objective is minimizing the overall energy cost of all buildings. We streamlined the state features by choosing the following key observations: \textit{outdoor dry bulb temperature $T_{t}^{\text{out}}$, outdoor relative humidity $H_{t}^{\text{out}}$, battery storage state $B_{t}^{\text{soc}}$, net electricity consumption $E_{t}^{\text{net}}$, electricity pricing $C_{t}$, and hour of the day $t$}. For the reward, we defined it as the negative net consumption from the grid for a building $E_{t,i}^{\text{grid}}$, where $E_{t,i}^{\text{load}}$ is building's non-shiftable load, and $E^{\text{solar}}_{t,i}$. The table \ref{tab:rl_case} presents an overview of the RL case.

\begin{table}[htb]
    \renewcommand*{\arraystretch}{1.1}
    \centering
    \begin{tabular}{@{}lc@{}}

        \toprule

        Action &
        $E^{\text{batt}}_{t,i} \in [-1, 1]$
        \\ \hdashline[.4pt/1pt]
        
        State &
        $\{T_{t}^{\text{out}}, H_{t}^{\text{out}}, B_{t}^{\text{soc}}, E_{t}^{\text{net}}, C_{t}, t\}$
        \\ \hdashline[.4pt/1pt]
        
        Reward &
        $-E_{t,i}^{\text{grid}} = -\max\{ E_{t,i}^{\text{load}} + E^{\text{batt}}_{t,i} - E^{\text{solar}}_{t,i}, 0\}$
        \\ \bottomrule
        
    \end{tabular}
    \vskip-5pt
    \caption{RL case definition}
    \label{tab:rl_case}
\end{table}
We represented the non-shiftable load and solar panel generation as a non-linear function of temperature, humidity, and some respective bases, reflecting real-world dynamics. That simulates real-world scenarios and effectively demonstrates the collaborative learning capacity of interrelations between features with FL. Additionally, for a personalized approach, we have set different coefficients for each building to cater to individual energy consumption patterns and characteristics, such as the air conditioner efficiency and the solar panel size. We present details of the environment in the Appendix.

\section{Method} \label{sec:model}
\begin{figure}[htb]
    \centering
    \includegraphics[width=0.7\linewidth]{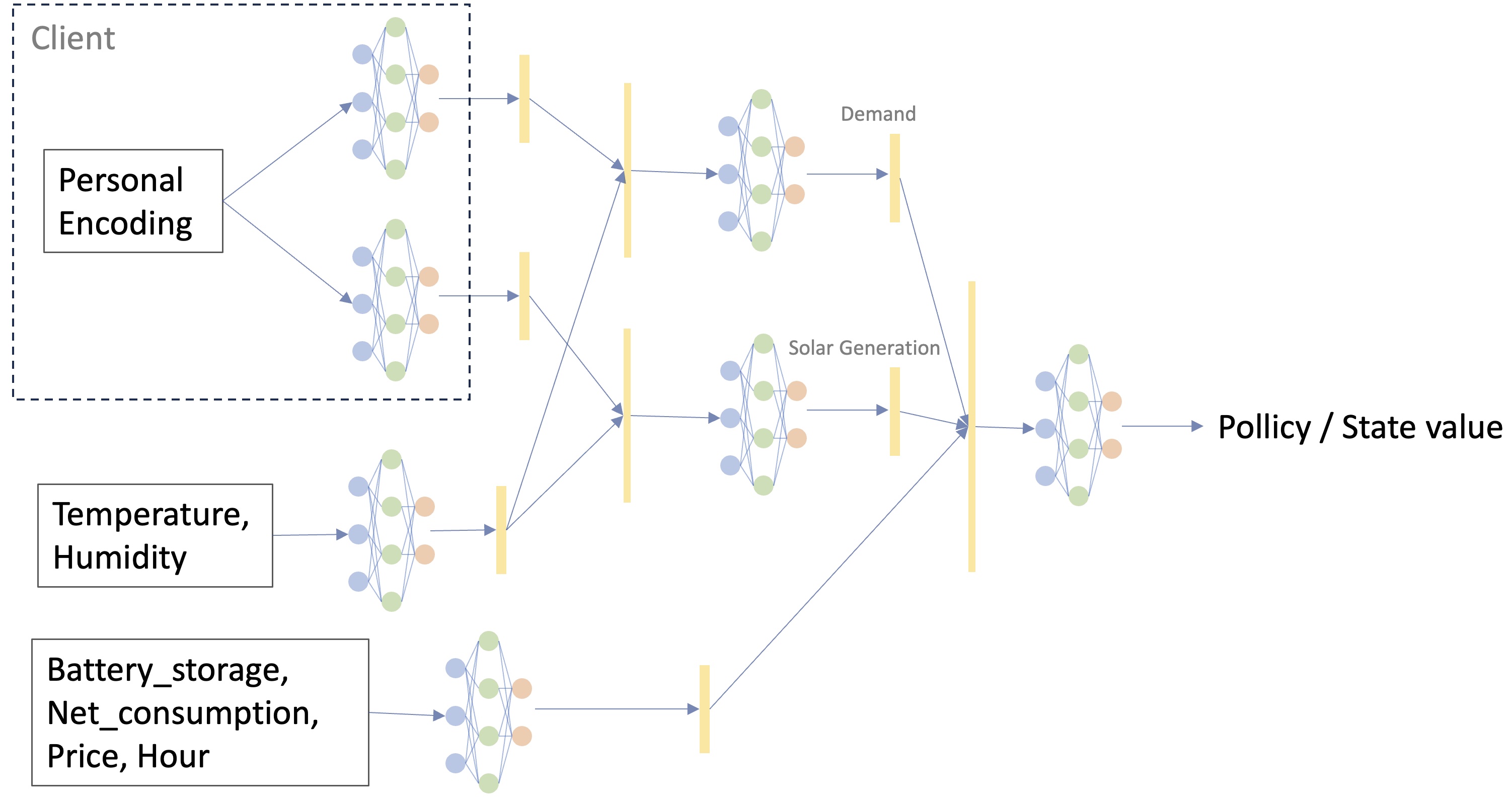}
    \caption{Our model captures the inherent interdependencies among features in mapping between states and policies.}
    \label{fig:model}
\end{figure}

{\bf The model.}
Understanding how features interact, we designed a model to capture these interactions faithfully, aiming to ensure the model's effectivity to represent the complex connections between features and the optimal policy. The architecture of our model is depicted in Figure \ref{fig:model}. We divided it into two parts: a personal component on the client side capturing individual information and a shared component capturing common features to enhance generalization. 
The personal component accommodates each building's unique demand and solar generation profiles, which are exclusively accessible to each building client, maintaining privacy in line with \cite{DBLP:journals/corr/abs-1812-00564}.
On the other hand, the shared component captures the shared information by concatenating the temperature, humidity, and personalized encoding, which embodies these distinctive patterns. Subsequently, we obtain the policy and state values for agent actions by concatenating this with the processed remaining features through an additional neural network layer.

{\bf FL TRPO with details.}
We use FL to collectively discern the interconnected relationships among demand, solar generation, temperature, and humidity across buildings. Relying solely on data from a single building risks overfitting, but leveraging diverse data distributions from multiple buildings helps uncover latent connections. Moreover, the model adeptly extrapolates unseen data instances by assimilating insights from disparate building data distributions, handling anomalies like households on vacation or harsh weather conditions.

In our problem, a key challenge arises with the possibility of the agent overcharging and discharging the battery, resulting in identical rewards as standard actions.
To address this, we introduce a penalty to discourage unwarranted overcharging and discharging.
Moreover, the agent has to make a sequence of pivotal decisions: charging during periods of low electricity pricing and discharging when prices are high.
Given the expansive search space, we used TRPO as it adeptly explores the vast space by identifying suitable step sizes within a confined region, ensuring stable policy, monotonic improvement, and more effective exploration than other policy gradient methods.

\section{Experiments}

We demonstrate the generalization ability of our model by employing different training and testing data distributions. The reader can find detailed information regarding data generation in Appendix \ref{sec:exp_detail}. Using testing data, we establish our baseline, denoted as \textit{Upperbound}, by training individual agents in each building. Then, we trained individual agents on each building using testing data as a reference that does not use FL, denoted as \textit{Ind. Agent}. We then compare the performance of FL TRPO against the conventional approach of training separate TRPO agents per building.

\begin{figure}[htb]
\centering
\begin{subfigure}{.5\textwidth}
  \centering
  \includegraphics[width=\linewidth]{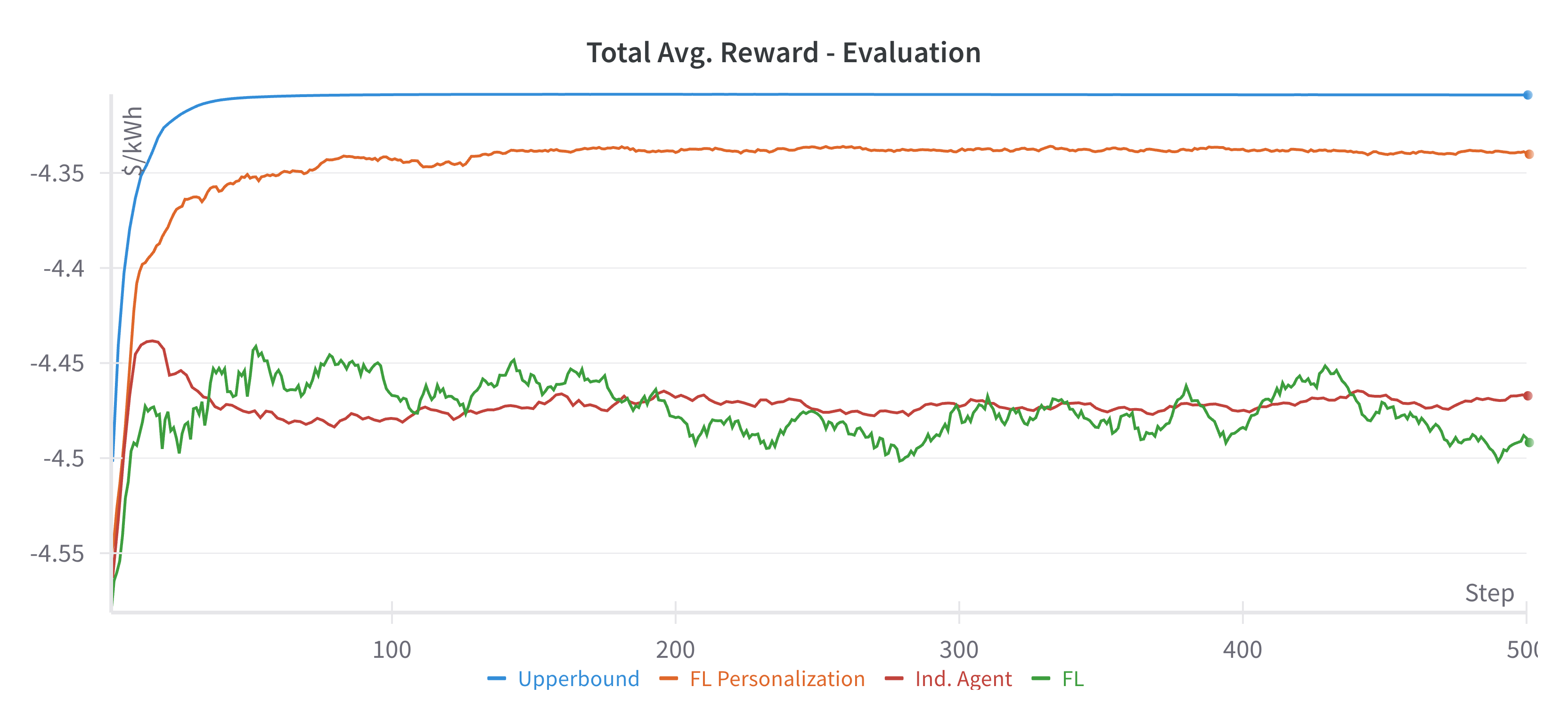}
  \label{fig:b1}
\end{subfigure}%
\begin{subfigure}{.5\textwidth}
  \centering
  \includegraphics[width=\linewidth]{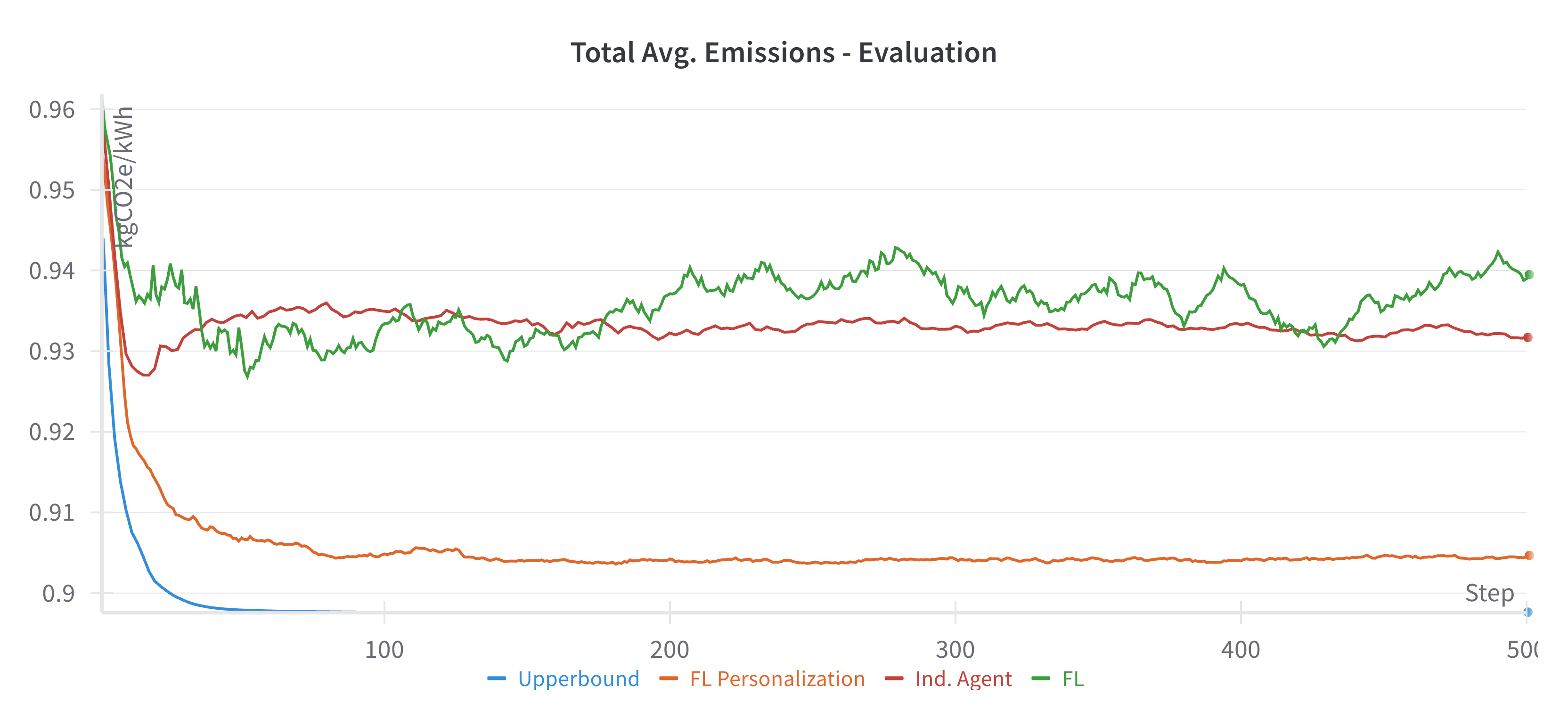}
  \label{fig:b2}
\end{subfigure}

\caption{Average reward and emission of the five buildings across five random seeds. \textbf{Upperboud (Blue):} A single TRPO agent trained using the testing dataset to establish the upper-performance limit. \textbf{FL (Green):} Model structured with all parts shared trained with FL methodology. \textbf{Ind. Agent (Red):} TRPO agent trained separately for each building. \textbf{FL Personalization (Orange):} FL TRPO with personalized encoding as detailed in Section \ref{sec:model}, trained using FL methodology.}
\label{fig:simple_results}
\end{figure}

\begin{figure}[htb]
\centering
\begin{subfigure}{.5\textwidth}
  \centering
  \includegraphics[width=\linewidth]{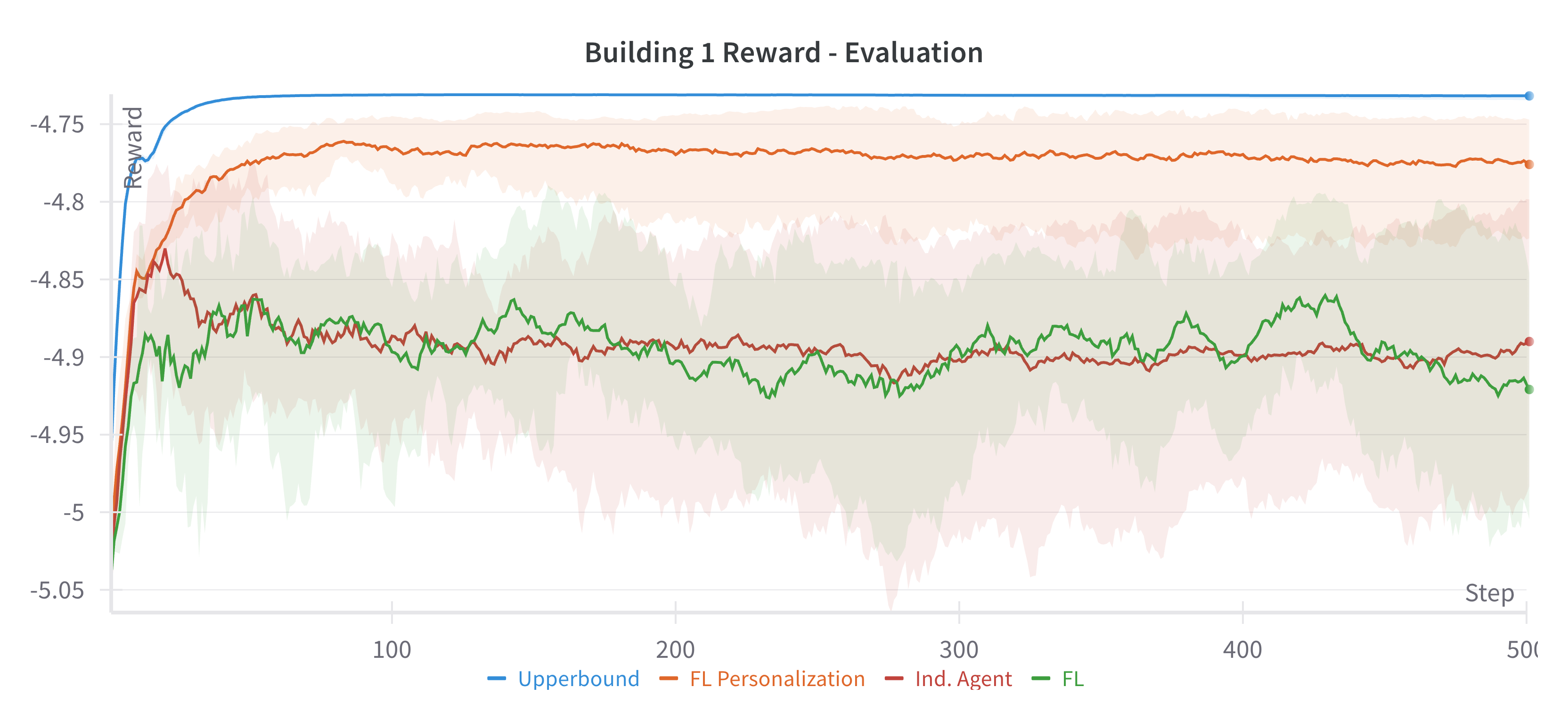}
  \label{fig:b1_r}
\end{subfigure}%
\begin{subfigure}{.5\textwidth}
  \centering
  \includegraphics[width=\linewidth]{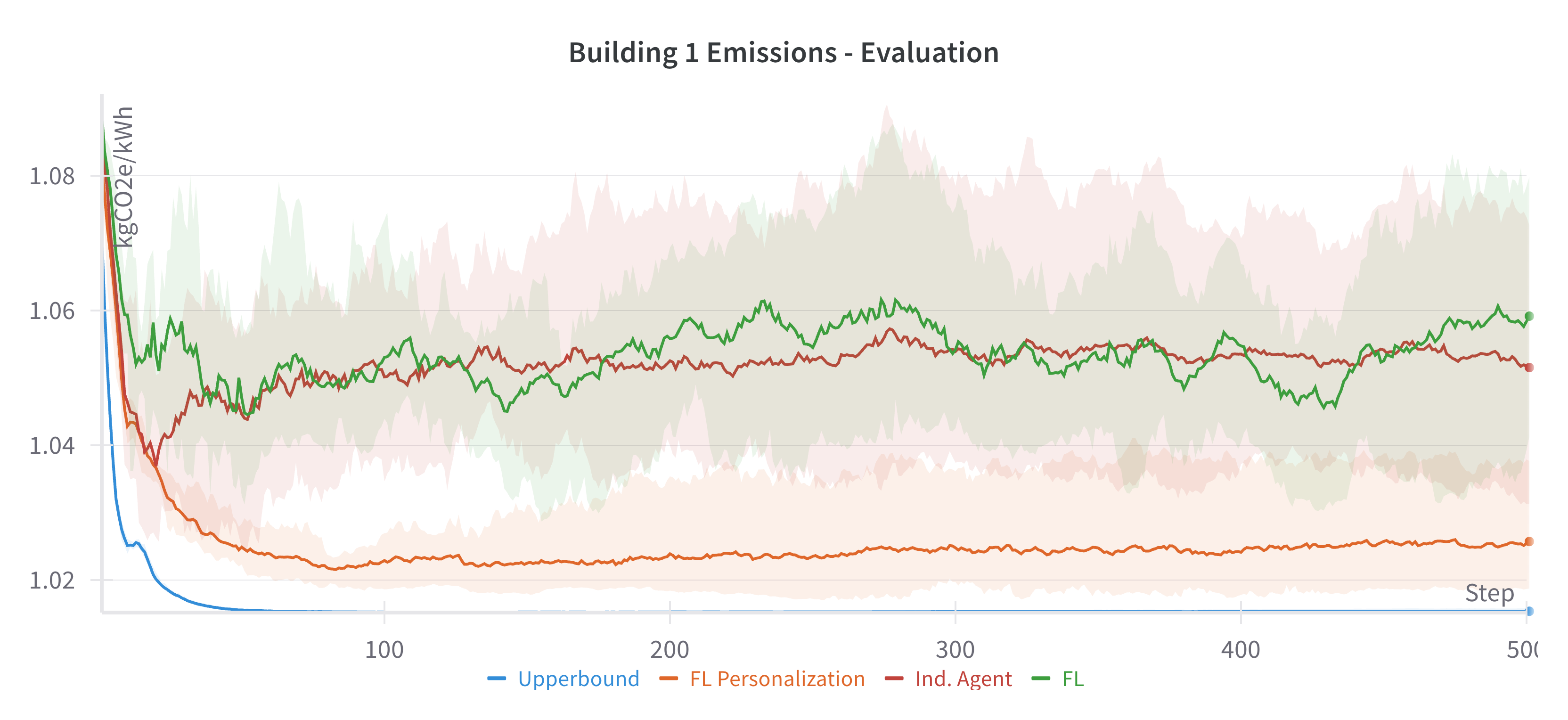}
  \label{fig:b1_e}
\end{subfigure}
\begin{subfigure}{.5\textwidth}
  \centering
  \includegraphics[width=\linewidth]{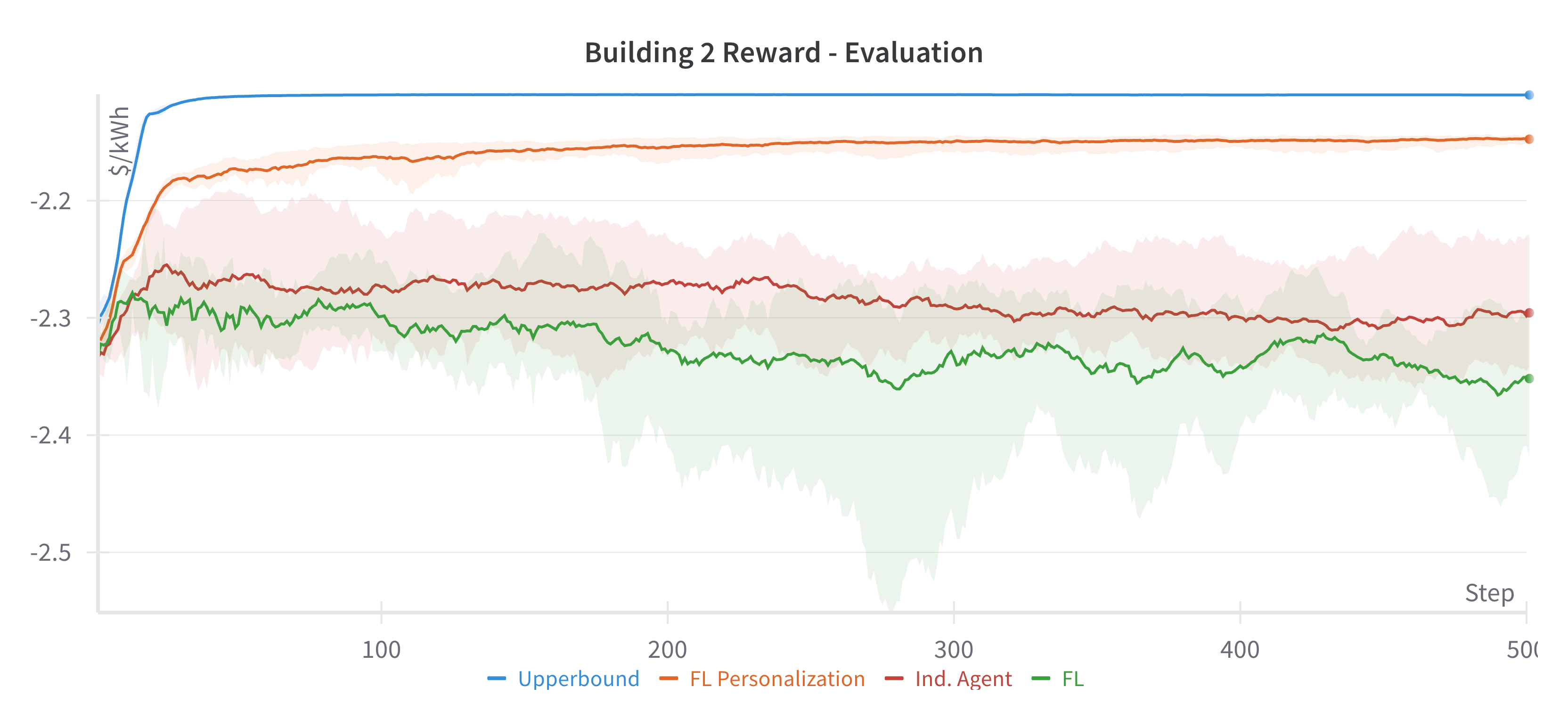}
  \label{fig:b2_r}
\end{subfigure}%
\begin{subfigure}{.5\textwidth}
  \centering
  \includegraphics[width=\linewidth]{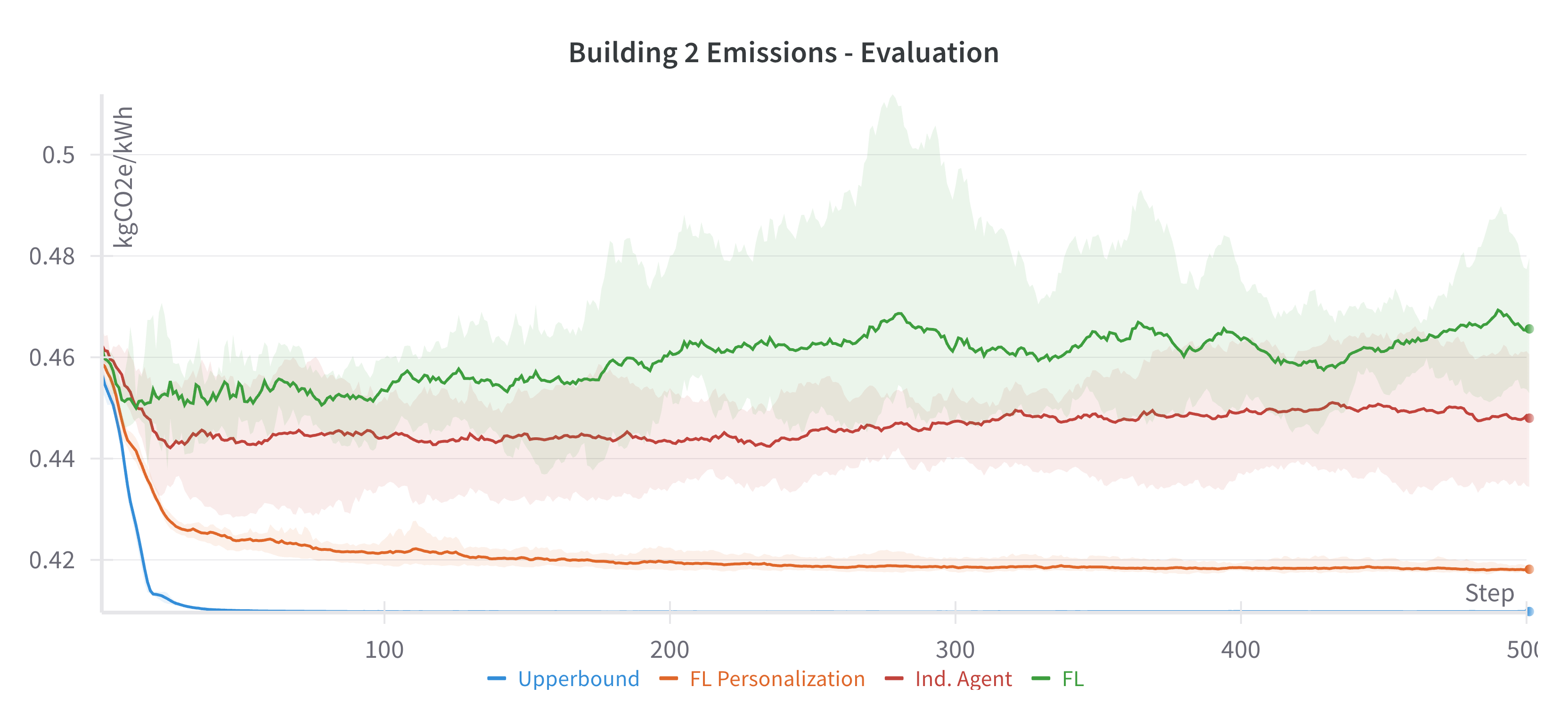}
  \label{fig:b2_e}
\end{subfigure}
\begin{subfigure}{.5\textwidth}
  \centering
  \includegraphics[width=\linewidth]{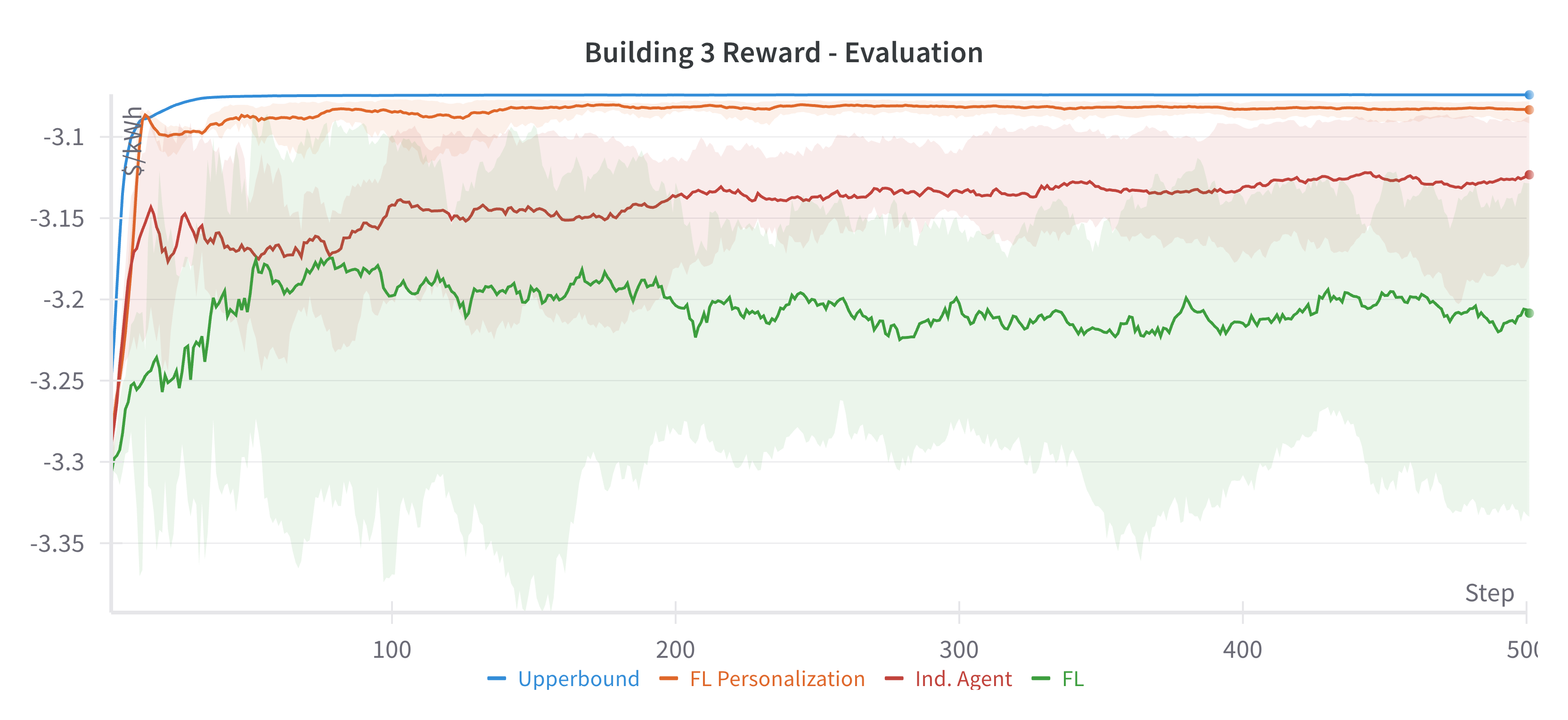}
  \label{fig:b3_r}
\end{subfigure}%
\begin{subfigure}{.5\textwidth}
  \centering
  \includegraphics[width=\linewidth]{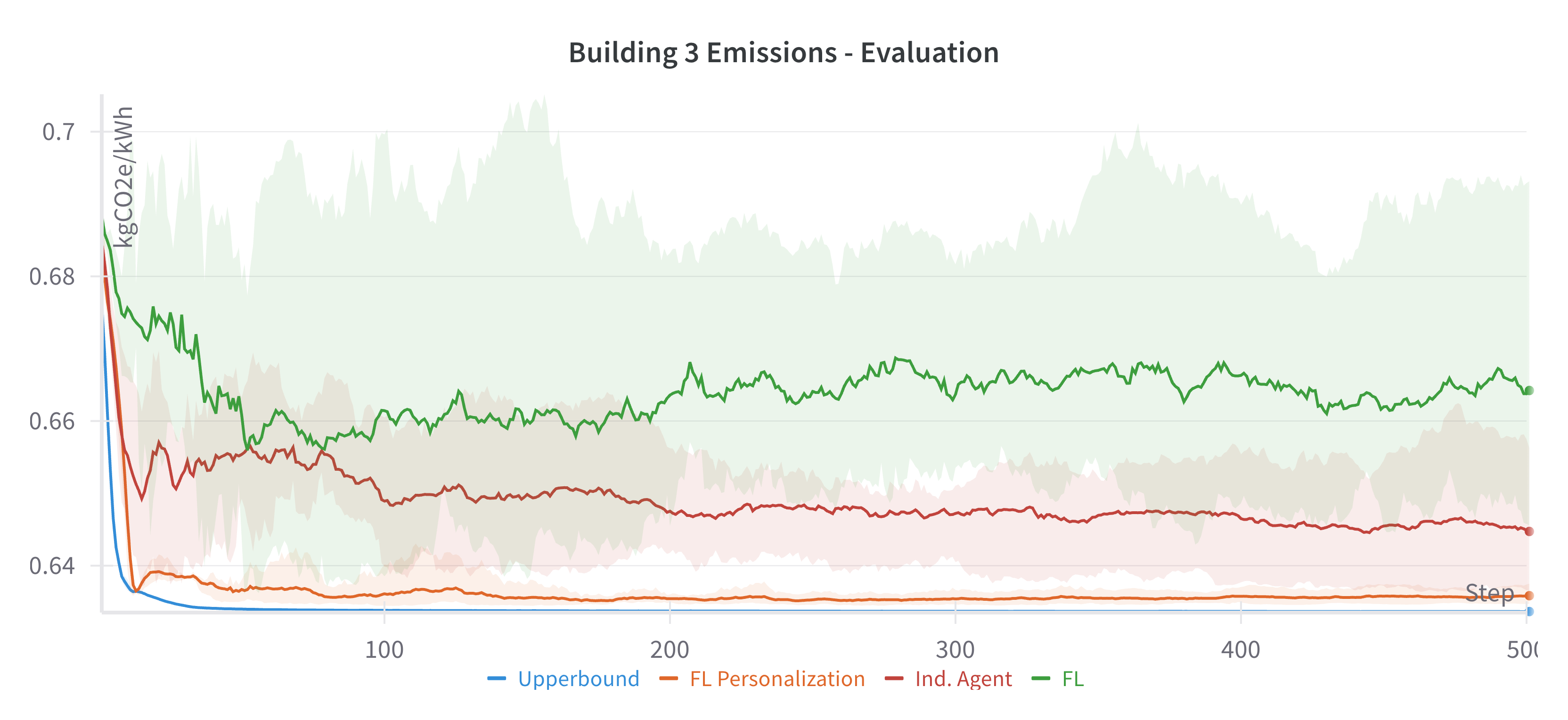}
  \label{fig:b3_e}
\end{subfigure}
\begin{subfigure}{.5\textwidth}
  \centering
  \includegraphics[width=\linewidth]{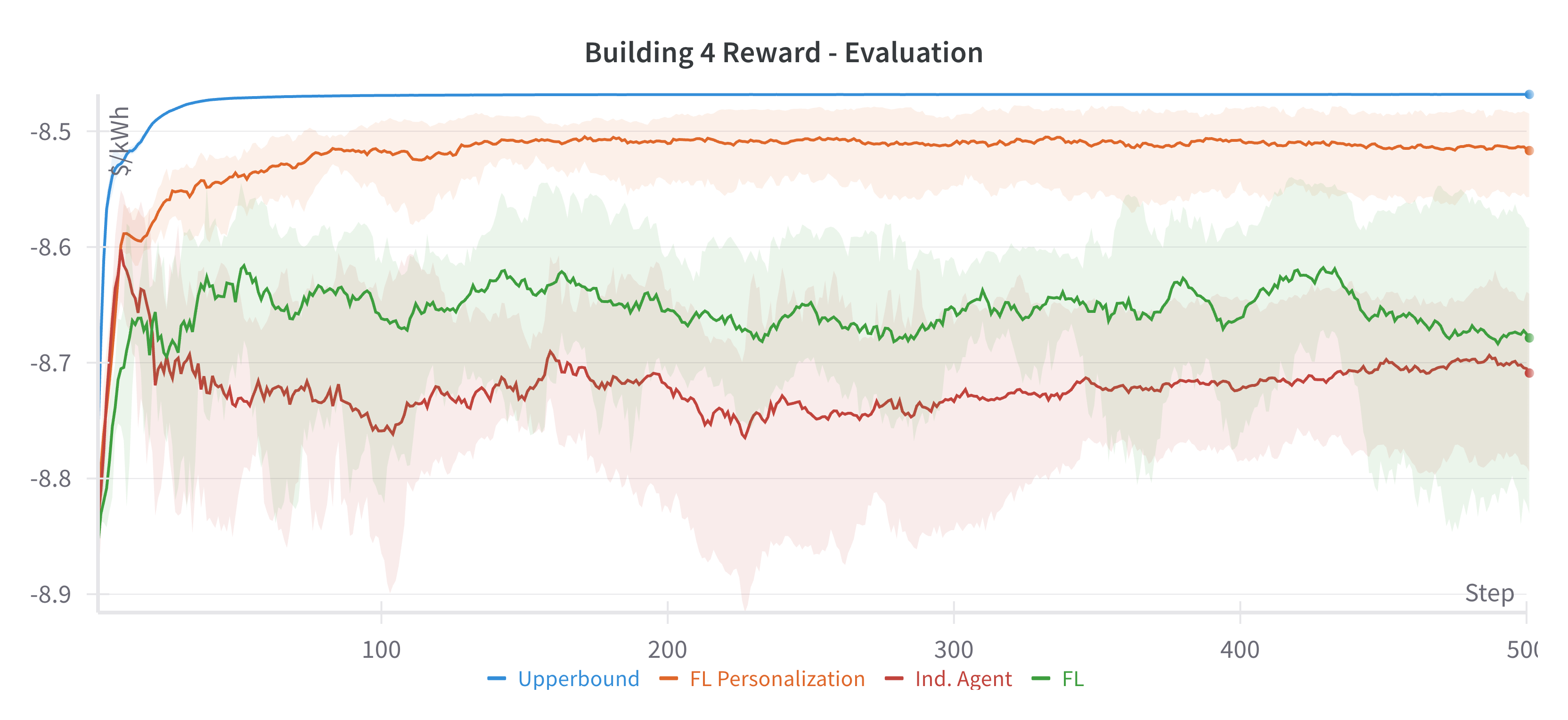}
  \label{fig:b4_r}
\end{subfigure}%
\begin{subfigure}{.5\textwidth}
  \centering
  \includegraphics[width=\linewidth]{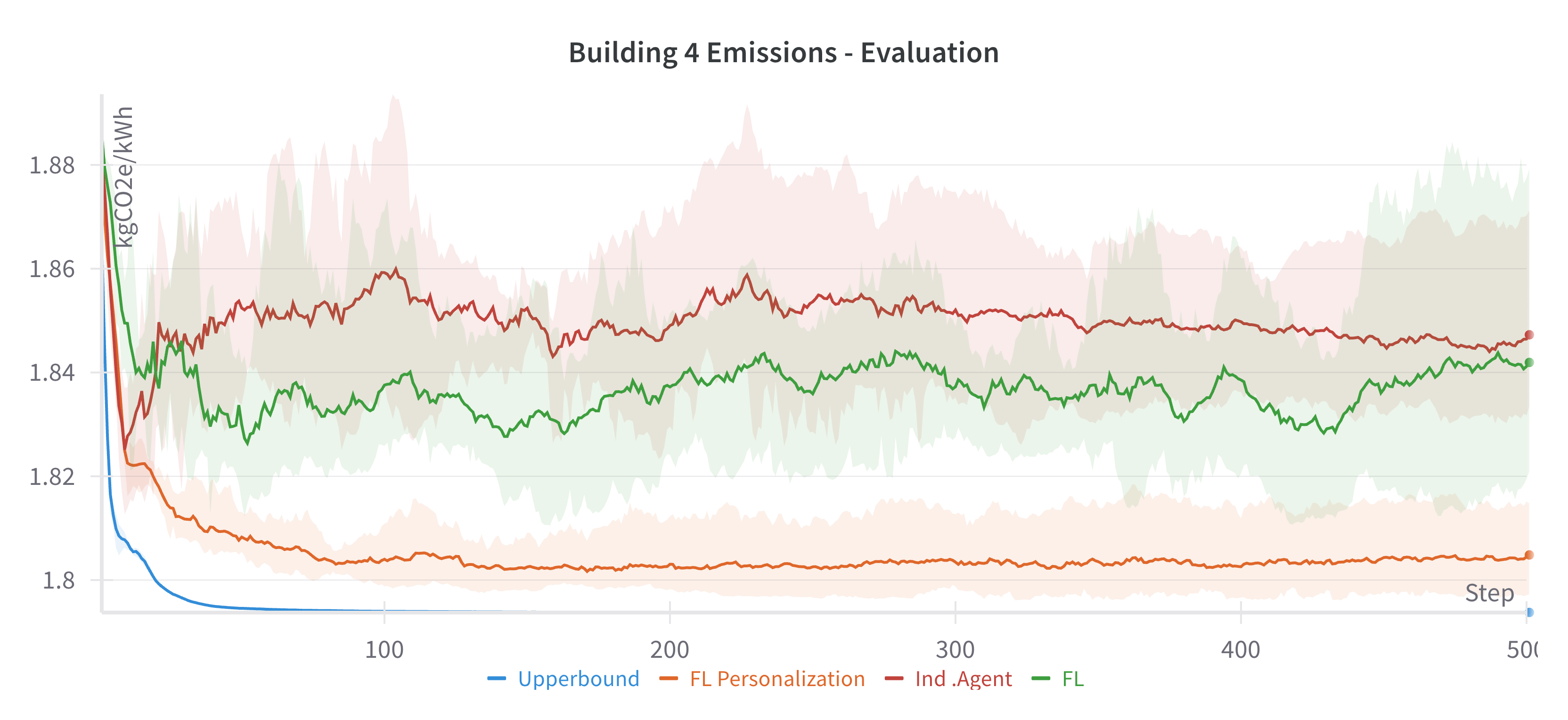}
  \label{fig:b4_e}
\end{subfigure}
\begin{subfigure}{.5\textwidth}
  \centering
  \includegraphics[width=\linewidth]{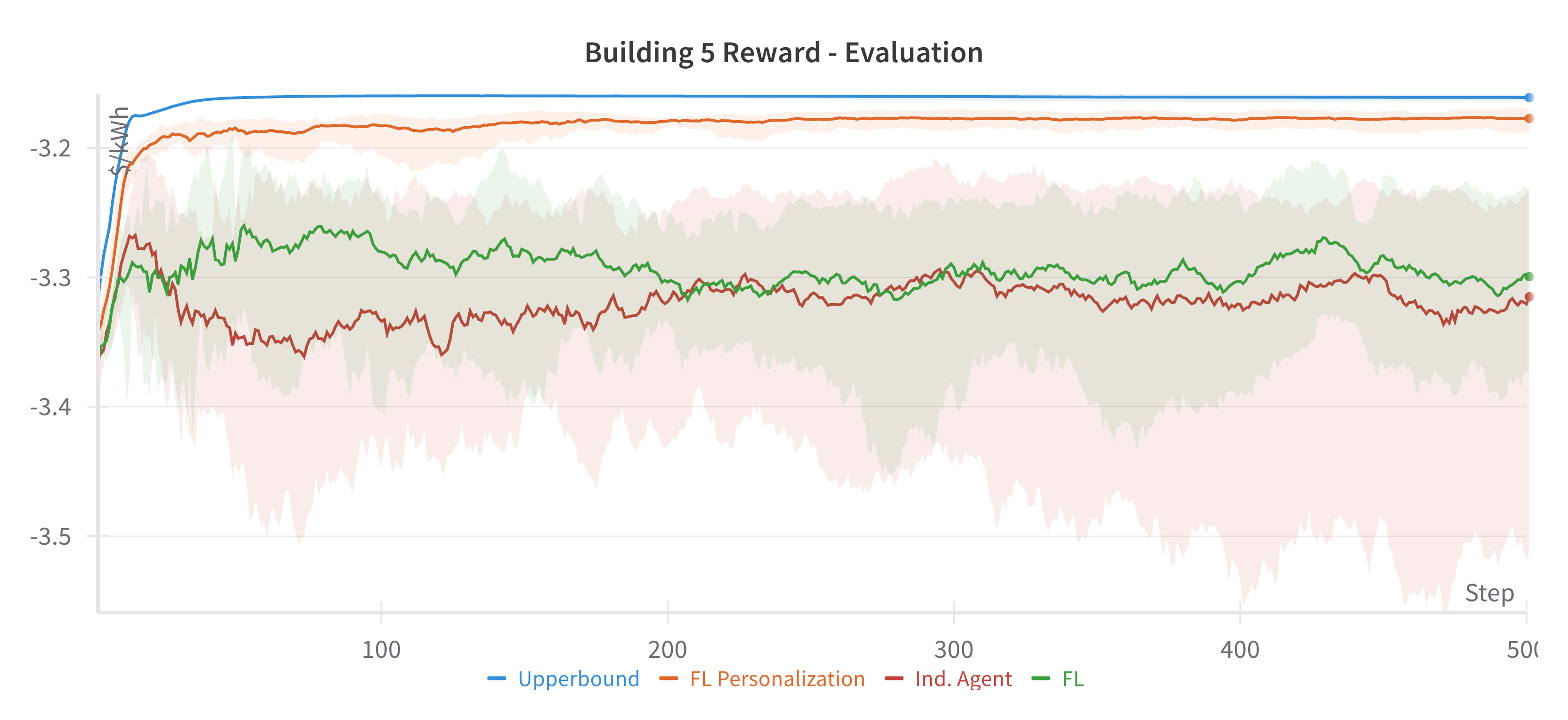}
  \label{fig:b5_r}
\end{subfigure}%
\begin{subfigure}{.5\textwidth}
  \centering
  \includegraphics[width=\linewidth]{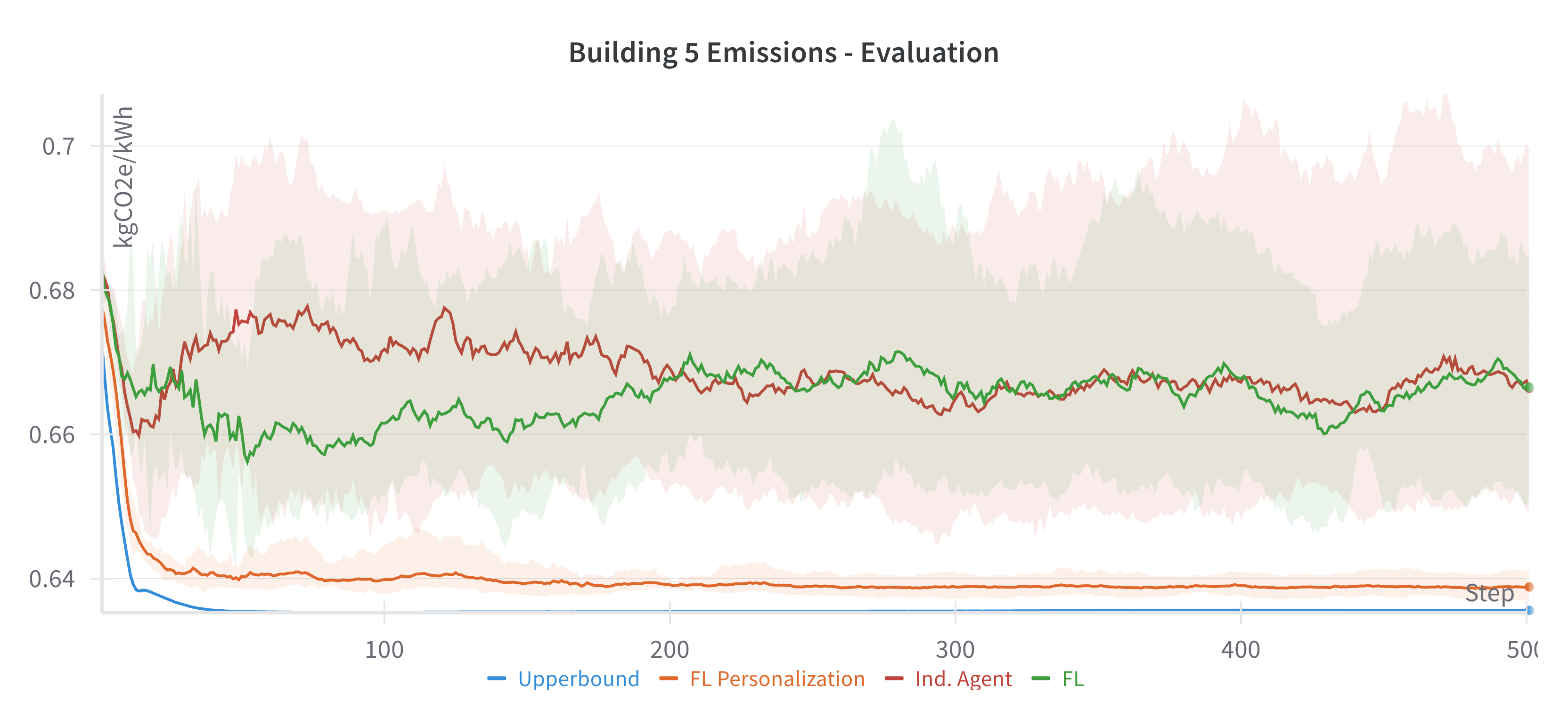}
  \label{fig:b5_e}
\end{subfigure}
\caption{Reward and emission of all buildings across five random seeds. \textbf{Upperboud (Blue):} A single TRPO agent trained using the testing dataset to establish the upper-performance limit. \textbf{FL (Green):} Model structured with all parts shared trained with FL methodology. \textbf{Ind. Agent (Red):} TRPO agent trained separately for each building. \textbf{FL Personalization (Orange):} FL TRPO with personalized encoding as detailed in Section \ref{sec:model}, trained using FL methodology. The performance of FL Personalization closely approaches the optimal baseline. We explain this improvement due to the integration of a model that effectively captures inherent feature relationships and personalized optimal policies.}
\label{fig:results}
\end{figure}

Additionally, we conduct an ablation study of the personalized encoding component by comparing our method with a model structured to share all features, denoted as \textit{FL}. Our model, referred to as \textit{FL Personalization}, incorporates personalized encoding. Figure \ref{fig:simple_results} presents five buildings' average reward and emission trends. The Appendix \ref{fig:results} contains further details regarding the reward and emission for each building.

As depicted in Figures \ref{fig:simple_results} and \ref{fig:results}, FL TRPO consistently outperforms \textit{FL} and \textit{Ind. Agents} across all buildings, approaching optimal baselines. Notably, \textit{FL} and \textit{Ind. Agents} exhibit similar performance, indicating that a simple model with FL training may not yield improvements. We hypothesize that different buildings require distinct optimal policies with the same state. \textit{FL}, lacking the ability to differentiate between buildings, may compromise performance. In contrast, with personalized encoding, FL leverages our prior knowledge, learns shared information, and can tailor policies for different clients.

\section{Conclusion}

This paper delved into applying FL to the TRPO approach in multi-building energy management. Our results underscore the effectiveness of FL in facilitating TRPO to discern intricate feature interdependencies, thereby enabling robust generalization across unobserved data distributions. Furthermore, introducing the integration of a feature-based model enhanced the performance of FL TRPO. This contribution validates the potential of our methodology to learn policies that effectively reduce both emissions and energy costs within microgrids while guaranteeing generalization and privacy preservation.

\clearpage
\bibliographystyle{plainnat}
\bibliography{ref}

\newpage

\section*{Appendix}

The source code and demo files have been anonymized and are available in \href{https://anonymous.4open.science/r/FRL-CE44/README.md}{\textbf{this repository}} link.

\subsection{Preliminaries}
\subsubsection{FedAvg}
FedAvg is a fundamental FL technique that trains a centralized model while respecting data decentralization and privacy; it initiates the process by training localized models using individual data. After a series of model updates, the parameters of these local models are transmitted to a central server and averaged. This iterative cycle of local training and central aggregation is recurrently executed by FedAvg, progressively enhancing the global model's efficacy with distinct datasets. Equation \ref{eq:fed_avg} presents the FedAvg update, in which the letter $\theta$ represents the parameters to train, $\eta$ a learning rate, $n_k$ the data samples of client $k$, $n$ the total number of samples and $g_k$ the gradient at client $k$.
\begin{equation}
    \theta_{t+1} = \theta_{t} - \eta \sum_{k=1}^K \frac{n_k}{n} g_k .
    \label{eq:fed_avg}
\end{equation}

\subsubsection{TRPO}
RL addresses sequential decision problems modeled as Markov Decision Processes (DMP). The problem is defined by the tuple ($\mathcal{S}$, $\mathcal{A}$, $\mathcal{P}$, $s$, $\gamma$), where $\mathcal{S}$ and $\mathcal{A}$ denote the observation and action spaces. The transition function $\mathcal{P}: \mathcal{S} \times \mathcal{A} \rightarrow \mathcal{S}$ represents state transitions' probabilities, and the reward function $r: \mathcal{S} \times \mathcal{A} \times \mathcal{S} \rightarrow \mathbb{R}$ provides immediate rewards for actions. The variable $\gamma \in [0, 1]$ denotes the discounted factor. A policy on an MDP is a mapping function $\mathcal{S} \rightarrow \mathcal{A}$. RL's primary objective is finding a policy that maximizes the cumulative rewards. Given a policy $\pi$, let $V^\pi$ denote the value function of state $s$ defined as
\begin{equation*}
    V^\pi(s) = \mathbb{E}_\pi \left[ \sum_{t=0}^\infty \gamma^t r_t \mid s_0 = s \right],
\end{equation*}
and let $Q^\pi$ denote the Q value for a state-action pair $(s, a)\in \mathcal{S}\times \mathcal{A}$
\begin{equation*}
    Q^\pi(s, a) = \mathbb{E}_\pi \left[ \sum_{t=0}^\infty \gamma^t r_t \mid s_0=s, a_0=a \right].
\end{equation*}

Then we can define the advantage function $A^\pi(s, a)$ as
\begin{equation*}
    A(s, a) = Q(s, a) - V(s).
\end{equation*}

Policy Gradient (PG) methods constitute a major branch among RL algorithms. While simple PG algorithms share similarities with first-order gradient methods, they are prone to overconfidence and a subsequent decline in performance. Trust Region Policy Optimization (TRPO) \cite{DBLP:conf/icml/SchulmanLAJM15} emerges as a cutting-edge PG algorithm to tackle this issue. TRPO takes substantial steps to ensure that improvements remain within predefined bounds. It does so by defining a surrogate advantage function $\mathcal{L}(\theta_k, \theta)$ that measures how much the new policy $\pi_{\theta}$ changed concerning the old policy $\pi_{\theta_k}$. The change is measured using an average KL-Divergence denoted $\hat{D}_{KL}(\theta_k||\theta)$. We framed the problem as a trust region optimization problem with a surrogate objective:
\begin{align}
    \mathcal{L}(\theta_k, \theta) &= \mathbb{E}_{s,a \sim \pi_{\theta_k}}\left[ \frac{\pi_\theta(a|s)}{\pi_{\theta_k}(a|s)} A^{\pi_{\theta_k}}(s,a)\right], \\
    \theta_{t+1} & = arg \max_\theta \mathcal{L}(\theta_k, \theta)~\text{s.t.}~\hat{D}_{KL}(\theta_k||\theta) \leq \gamma.
\end{align}
Its comprehensive search for the optimal direction and step length enables TRPO to require fewer update steps than standard policy gradient algorithms. Consequently, TRPO demands less gradient communication in FL algorithms, making it an ideal choice for FL, which relies on efficient communication.

\subsection*{Environment details} \label{app:env_details}
\begin{wrapfigure}{r}{0.4\textwidth}
    \centering
        \vskip-50pt
    \includegraphics[width=0.4\textwidth]{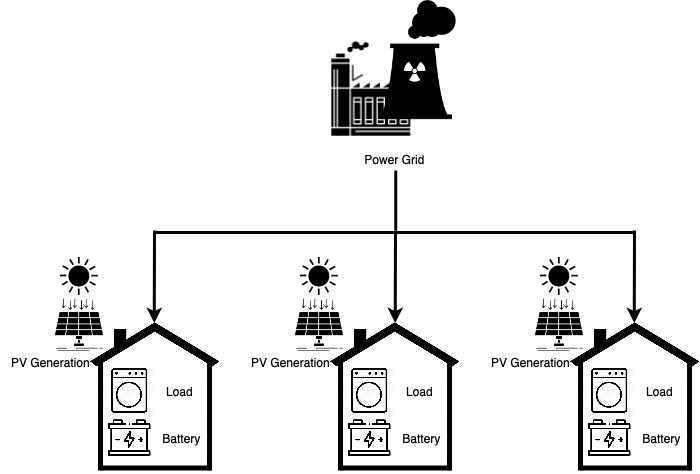}
    \caption{Scenario definition.}
    \label{fig:problem}
        \vskip-20pt
\end{wrapfigure}
Each building features a photovoltaic panel, a non-shiftable load, and a battery. The buildings can procure electricity from the grid, store energy generated by the photovoltaic panels, and regulate the battery by charging and discharging to fulfill the required load.

\subsection{Experiment settings} \label{sec:exp_detail}

We compared FL TRPO's performance to training one agent per building to demonstrate that our algorithm benefits from unseen data distribution scenarios. Without collaboration or information sharing, one agent per building is the same as decentralized multi-agent reinforcement learning. 

\begin{figure} [hbt]
\centering
\begin{subfigure}{.3\textwidth}
  \centering
  \includegraphics[width=\linewidth]{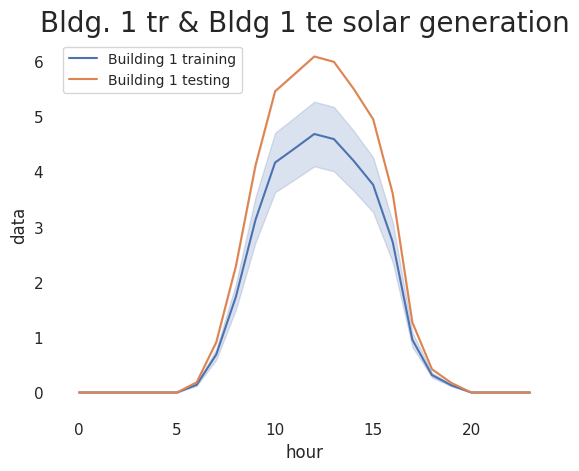}
  \label{fig:solar_11}
\end{subfigure}%
\begin{subfigure}{.3\textwidth}
  \centering
  \includegraphics[width=\linewidth]{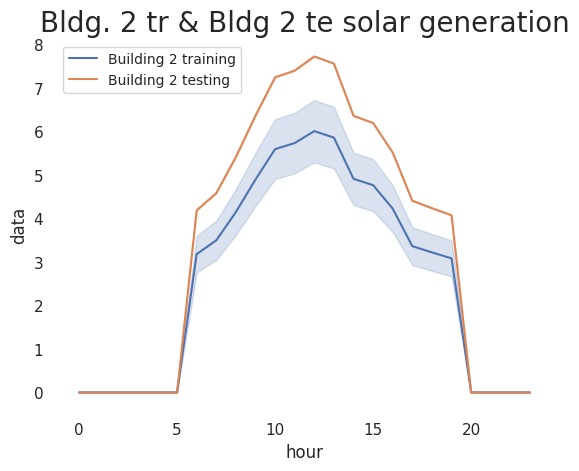}
  \label{fig:solar_22}
\end{subfigure}
\begin{subfigure}{.3\textwidth}
  \centering
  \includegraphics[width=\linewidth]{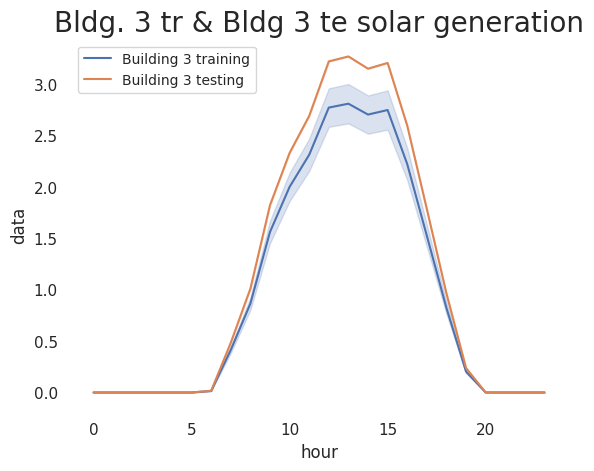}
  \label{fig:solar_33}
\end{subfigure}
\begin{subfigure}{.3\textwidth}
  \centering
  \includegraphics[width=\linewidth]{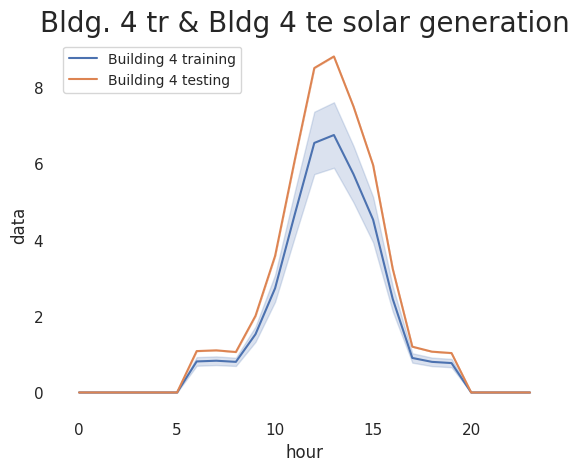}
  \label{fig:solar_44}
\end{subfigure}
\begin{subfigure}{.3\textwidth}
  \centering
  \includegraphics[width=\linewidth]{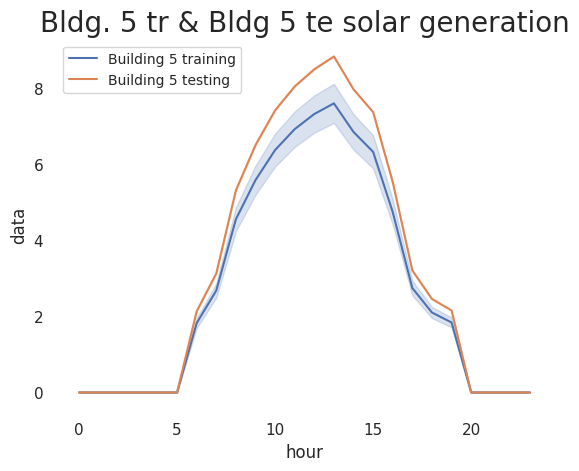}
  \label{fig:solar_55}
\end{subfigure}%
\caption{Training and testing solar generation data of each building.}
\label{fig:solar_single}
\end{figure}

\begin{figure}[htb]
\centering
\begin{subfigure}{.3\textwidth}
  \centering
  \includegraphics[width=\linewidth]{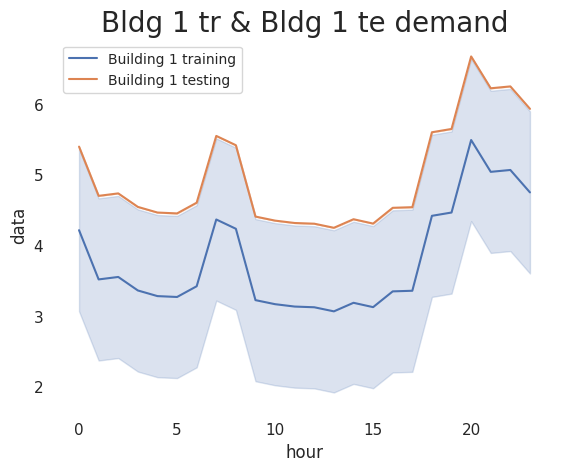}
  \label{fig:demand_11}
\end{subfigure}%
\begin{subfigure}{.3\textwidth}
  \centering
  \includegraphics[width=\linewidth]{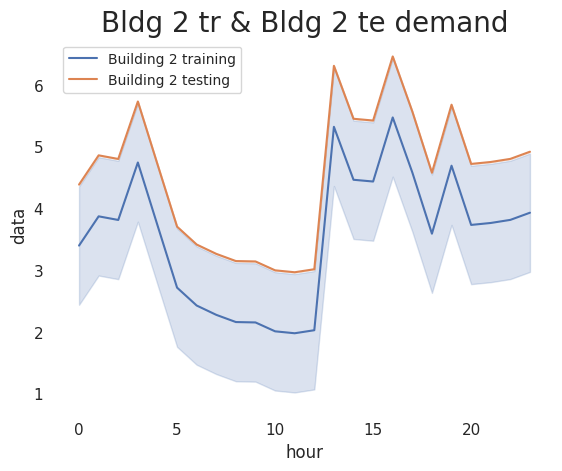}
  \label{fig:demand_22}
\end{subfigure}
\begin{subfigure}{.3\textwidth}
  \centering
  \includegraphics[width=\linewidth]{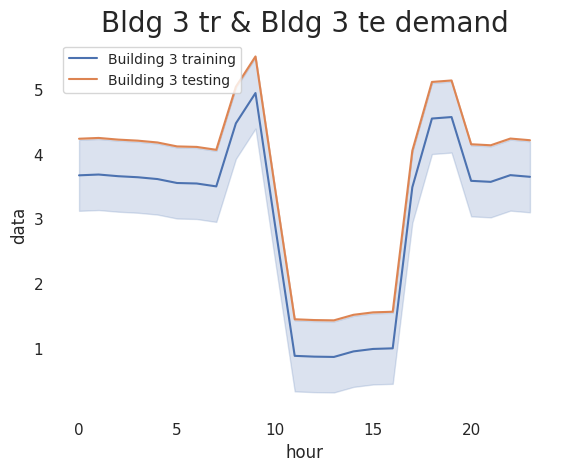}
  \label{fig:demand_33}
\end{subfigure}%
\begin{subfigure}{.3\textwidth}
  \centering
  \includegraphics[width=\linewidth]{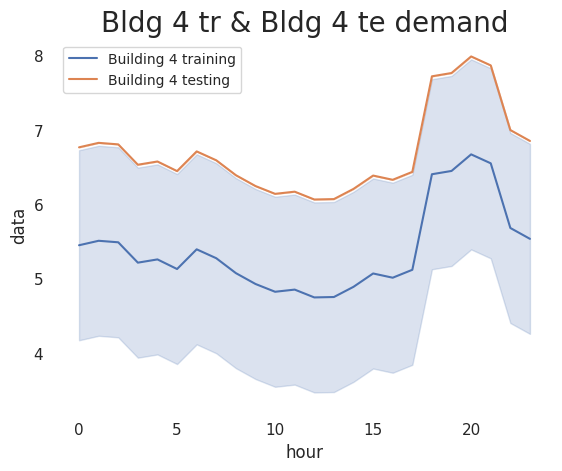}
  \label{fig:demand_44}
\end{subfigure}
\begin{subfigure}{.3\textwidth}
  \centering
  \includegraphics[width=\linewidth]{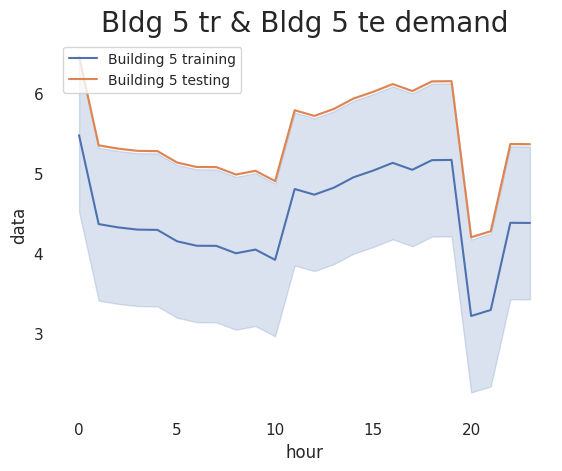}
  \label{fig:demand_55}
\end{subfigure}%
\caption{Training and testing non-shiftable load data of each building.}
\label{fig:demand_single}
\end{figure}

Non-shiftable load and solar generation depend on non-linear functions that rely on environmental factors like temperature, humidity, and building-specific base patterns.
Furthermore, to substantiate the efficacy of our approach in generalizing the ability to unseen data distributions, we introduce random variables with distinct ranges to the temperature and humidity data generation during the training and testing phases. We randomly sample noise from different ranges for each episode's data to assess our model's generalization ability, creating a distinction between training and testing data. This approach allows us to showcase the model's generalization capability across varying conditions.
In our experimental environment, we used five buildings characterized by unique configurations. Figure~\ref{fig:solar_single} and Figure~\ref{fig:demand_single} detail the training and testing range for solar generation and the non-shiftable load of the five buildings, correspondingly. 

\begin{figure}[htb]
\centering
\begin{subfigure}{.5\textwidth}
  \centering
  \includegraphics[width=\linewidth]{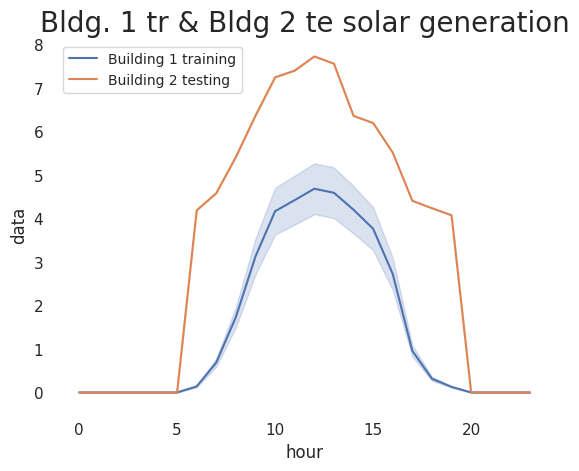}
  \label{fig:solar_12}
\end{subfigure}%
\begin{subfigure}{.5\textwidth}
  \centering
  \includegraphics[width=\linewidth]{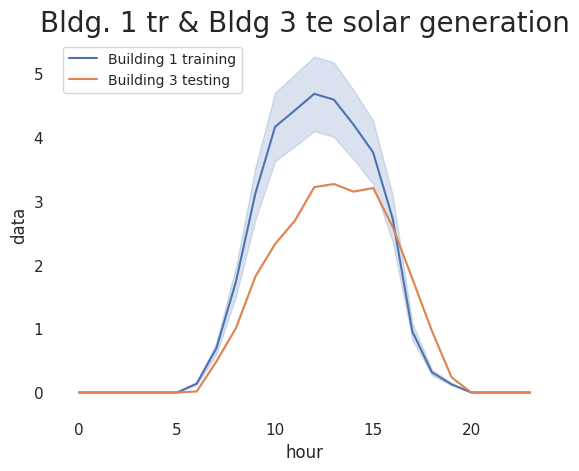}
  \label{fig:solar_13}
\end{subfigure}
\begin{subfigure}{.5\textwidth}
  \centering
  \includegraphics[width=\linewidth]{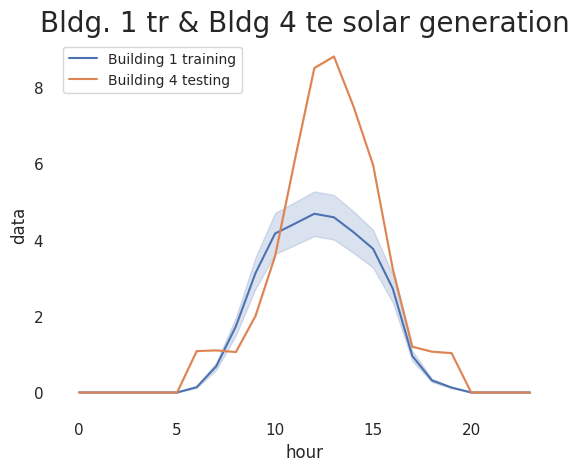}
  \label{fig:solar_14}
\end{subfigure}%
\begin{subfigure}{.5\textwidth}
  \centering
  \includegraphics[width=\linewidth]{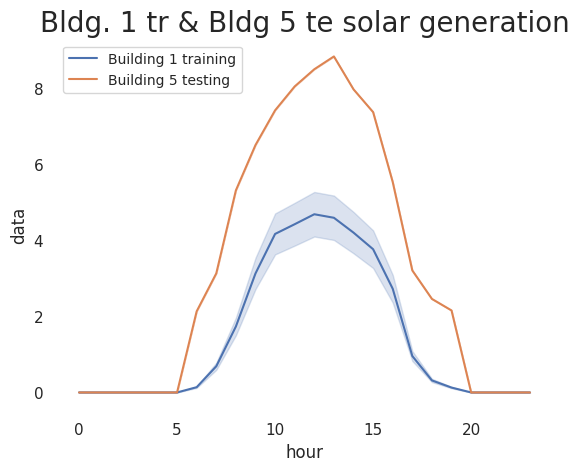}
  \label{fig:solar_15}
\end{subfigure}
\caption{Comparison between solar generation data of building one and the rest of buildings. FL can learn the hidden relationship among the features with different data distributions.}
\label{fig:comparison}
\end{figure}

During training and testing, we randomly sampled one set for each episode. As depicted in the figures, a disparity exists between the training and testing datasets. This partition shows the model's ability to generalize to unseen data distributions. To further illustrate our environment, we provide solar generation data for building one against other buildings in Figure \ref{fig:comparison}. With heterogeneous data distribution, our method can learn the hidden correlations and harness them for effective generalization.

\end{document}